\pdfoutput=1

\documentclass[11pt]{article}

\usepackage[]{EACL2023}

\usepackage{times}
\usepackage{latexsym}

\usepackage[T1]{fontenc}
\usepackage[utf8]{inputenc}

\usepackage{microtype}

\usepackage{inconsolata}

\usepackage{graphicx}
\usepackage{subcaption}
\newcommand{\ediv}{Pragmatically Appropriate Diversity}
\newcommand{\edivshort}{PA Diversity}
\newcommand{\edivspace}{Pragmatically Appropriate Diversity }
\newcommand{\edivshortspace}{PA Diversity }
\usepackage{multirow}
\usepackage{multicol}
\usepackage{enumitem}
\usepackage{longtable}

\title{\edivspace for Dialogue Evaluation}

\author{Katherine Stasaski \\
  Salesforce AI Research\thanks{~~Research performed  while at UC Berkeley.}\\
  \texttt{katie\_stasaski@berkeley.edu}\\\And
  Marti A. Hearst \\
  UC Berkeley \\
  \texttt{hearst@berkeley.edu}}

\begin{document}
\maketitle
\begin{abstract}
Linguistic pragmatics state that a conversation's underlying speech acts can constrain the type of response which is  appropriate at each turn in the conversation. 
When generating dialogue responses, neural dialogue agents struggle to produce \textit{diverse} responses.  Currently, dialogue diversity is assessed 
using automatic metrics, but the underlying speech acts do not  inform these  metrics.  

To remedy this, we propose the notion of \ediv, defined as the extent to which a conversation creates and constrains the creation of multiple diverse responses. 
Using a human-created multi-response dataset, we 
find significant support for the hypothesis that speech acts provide a signal for the diversity of the set of next responses.
Building on this result, we propose a new human evaluation task where creative writers predict the extent to which conversations inspire the creation of multiple diverse responses.  Our studies find  that   writers' judgments align with the \edivspace of  conversations.
Our work suggests that expectations for diversity metric scores should vary depending on the speech act.

\end{abstract}

\section{Introduction}

While many different   utterances can 
continue a conversation, some types of responses are more appropriate than others.
This work begins with the assumption, based on the linguistic pragmatics literature, that the speech acts associated with a conversation can constrain the responses which are  appropriate.  In particular, some speech acts are  part of \textit{adjacency pairs}, where a speaker produces a speech act utterance which is expected to be immediately followed in the conversation by the other speaker's paired speech act utterance \cite{levinson_1983}. For example, a Closing utterance by one speaker (such as ``Goodbye'') is typically followed by another Closing utterance from the other speaker \cite{schegloff}.  

Speech acts can  constrain a conversation even if not part of a strict adjacency pair. 
In relation to how speech acts influence conversational structure,
\citet{searle1985foundations} state: \begin{quote}
    The key to understanding the structure of conversations is to see that each illocutionary act creates the possibility of a finite and usually quite limited set of appropriate illocutionary acts as replies.  Sometimes the appropriate illocutionary act reply is very tightly constrained by the act that precedes it, as in question and answer sequences; and sometimes it is more open, as in causal conversations that move from one topic to another ... each illocutionary act in a conversation creates and constrains the range of appropriate illocutionary responses.
\end{quote}   
 
\begin{figure}[t]
    \centering
    \small
    \begin{tabular}{p{7.5cm}}
        \multicolumn{1}{c}{\bfseries Conversation A}\\
        \textit{Speaker 1:} Are you free tonight?\\
        \\
        \multicolumn{1}{c}{\bfseries Writer-Generated Responses}\\
        $\bullet$ No, not this evening. Can we try for tomorrow night?\\
        $\bullet$ Let's see, I'm free around 8pm. Will that work?\\
        $\bullet$ Yes! What are we getting into?\\
        $\bullet$ Well that depends on what you have planned.\\
        $\bullet$ Didn't we talk about this already? I have a work event tonight.
        \\
        \multicolumn{1}{c}{\bfseries Conversation B}\\
        \textit{Speaker 1:} Please come in and sit down. I’m happy to finally meet you. \\
        \hangindent=0.7cm \hspace{0.7cm}\textit{Speaker 2:} Same here, Ms. Drake. I've been looking forward to this. \\
        \textit{Speaker 1:} I’m sorry I kept you waiting.\\
        \\
        \multicolumn{1}{c}{\bfseries Writer-Generated Responses}\\
        $\bullet$ no problem at all!\\
        $\bullet$ no worries. how are you?\\
        $\bullet$ I was just reviewing some files so it's not a problem.\\
        $\bullet$ you are just in time!\\
        $\bullet$ Let's get started.\\
    \end{tabular}
    \caption{Two conversations from DailyDialog++ with sets of responses from writers (Section \ref{sec:writers}).  When considering different ways the speaker could continue the conversation, the \edivspace is higher for Conversation A (Yes-No Question) than for Conversation B (Apology). } 
    \label{fig:sample_convo}
\end{figure}

Consider the two conversations presented in Figure \ref{fig:sample_convo} from DailyDialog++, a conversational dataset with multiple human-provided responses \cite{sai-etal-2020-improving, li-etal-2017-dailydialog}.  We would expect the linguistic diversity of a set of responses for Conversation A to be higher than Conversation B because Conversation B contains a constraining utterance, an apology, as the most recent turn in the conversation. 
Speech act constraints dictate there are a limited number of ways to respond to an apology  utterance and remain pragmatically appropriate.  In fact, the first four writer-generated responses to this conversation include some form of Acceptance or Minimization.  By contrast, Conversation A ends with a question.  Although the question constrains the next response to contain an answer, the answer's content can be expressed in many different ways. 

Relevancy metrics and perplexity are common ways to measure quality and fluency of model-generated dialogue responses, respectively. 
Additionally, because models over-generate typical responses, another important evaluation criterion  is the  ability to produce \textit{diverse} responses, which capture multiple distinct, interesting ways a conversation can be continued.
Diversity  can be
measured using a variety of automatic metrics \cite{li-etal-2016-diversity,nli-placeholder,tevet-berant-2021-evaluating,reimers-gurevych-2020-making, cao-clark-2017-latent, zhu-et-al-acm, larson-etal-2019-outlier}.
However, these metrics do not vary given the conversation's properties; 
a model's output would be evaluated the same way whether in response to either conversation from Figure \ref{fig:sample_convo}. 
Speech act theory suggests that next turns in conversations are constrained, but 
currently, evaluations of chitchat dialogue models do not incorporate these constraints explicitly.  Our work explores whether automatic linguistic diversity metrics capture pragmatic constraints on appropriate responses.

To address this, we introduce the notion of \edivspace (\edivshort), defined as the extent to which a conversation's speech acts create and constrain the creation of multiple diverse responses.  
To explore this concept systematically, we conduct two analyses.  
First, we
examine both human-labeled and automatically-labeled speech acts of a human-generated multi-response conversational dataset. 
We hypothesize that the most recent speech act utterance constrains the diversity of responses.
We discover that automated diversity metrics, particularly NLI Diversity \cite{nli-placeholder}, detect differences among the most-recent speech act utterances.

We next propose a novel human evaluation task: rating the extent to which a conversational prompt inspires the creation of multiple diverse responses.  We choose to use creative writers as participants as past work 
finds that experts are reliably able to judge the quality of 
creative products while non-experts are not able to do this \cite{amabile, baer}. 
Our study finds significant differences among writers' \edivshortspace ratings for different speech acts.  We also find that these differences align with both the measured differences from the original dataset and our hypotheses about which speech acts cause more diverse responses.  To our knowledge, our findings are the first to combine speech acts with diversity-based assessment of neural conversational dialogue systems.

Our contributions in this paper are:
\begin{itemize}[noitemsep,topsep=0pt]
    \item The notion of \edivshort, the extent a conversation's speech acts creates and constrains creation of multiple diverse responses, 
    \item An automatic analysis with manually-assigned and automatically-labeled speech acts, finding that the automatic NLI Diversity metric varies based on the most recent speech act and align with our hypotheses, and
    \item A novel human-centered study, finding that expert creative writer judgments of PA Diversity align with NLI Diversity values.
\end{itemize}

\begin{table*}[ht!]
\small
    \centering
    \begin{tabular}{|p{4cm}|p{10.5cm}|}
     \multicolumn{2}{c}{\textbf{DailyDialog++ Speech Act Utterance Definitions and Examples}} \\\hline
     \rule{0pt}{1em}
     \textbf{Inform} & \textit{Definition:} ``the
speaker provides the addressee certain information which
he believes the addressee not to know or not to be aware
of, and which he assumes to be correct'' \\
& \textit{Example:} I could not relocate in the next year, but might be open to it in the future.\\\cline{2-2}
     \textbf{Question} & \textit{Definition:} ``the speaker wants to know something, which
he assumes the addressee to know, and puts pressure on the
addressee to provide this information'' \\
& \textit{Example:} Is there any extra pay for that?\\\cline{2-2}
     \textbf{Directive} & \textit{Definition:} ``concerned with the speaker’s wish
that the addressee performs an action''\\
     & \textit{Example:} Oh, thank you . I want one.\\\cline{2-2}
     \textbf{Commissive} & \textit{Definition:} `capture the speaker’s commitments to
perform certain actions'' \\
& \textit{Example:} You are probably right . I'll go right now and apologize . I try hard not to be late but it is difficult with Beijing traffic.\\\hline
     \multicolumn{2}{c}{}\\
     \multicolumn{2}{c}{\textbf{SWBD Speech Act Utterance Examples (Selected Subset)}}\\\hline
      \rule{0pt}{1em}\textbf{Statement Non Opinion}   & Actually, Costa Rica isn't in South America . It's in Central America.\\
     \textbf{Yes-No Question}   & Do you mind if I put my jacket there?\\
     \textbf{Wh Question}    & What materials do they need in order to apply the passport?\\
     \textbf{Action Directive}   & Let me see . \$200 makes RIB 1,616 . Here is the cash and exchange memo . Please check it.\\
     \textbf{Statement Opinion}    & Well, a good teacher makes good students.\\
     \textbf{Conventional Closing}    & Ok . Goodbye.\\
     \textbf{Open Question}   & How do you feel about it?\\
     \textbf{Offers, Options, Commits}    & Well, I'll call you immediately for instructions on the matter.\\
     \textbf{Thanking}    & Thank you for your encouragement.\\
     \textbf{Apology}    & Oh, I am so sorry.\\\hline
    \end{tabular}
    \caption{Speech acts from DailyDialog and SWBD along with a sample  conversational utterance from DailyDialog++.  DailyDialog++ speech act definitions are quoted from \citet{amanova-etal-2016-creating}.  For brevity, we only include one turn for each speech act instead of the entire conversation (full conversations can be seen in Appendix \ref{app:speechact_convos}).}
    \label{tab:speechact_table}
\end{table*}

\section{Related Work}\label{sec:related}
We place this work in the context of dialogue diversity, pragmatics, and creativity evaluation.

\subsection{Diversity Metrics for Dialogue}
Several automatic metrics have been proposed to measure the diversity of dialogue model responses. One category of metrics assumes the dialogue model generates \textit{one} response for each conversation in the test set (the \textit{Test Set Diversity} setting).  A frequently used metric is distinct-n, which measures the number of distinct n-grams across  all generated sequences \cite{li-etal-2016-diversity}.  Other metrics measure the number of distinct responses over the test set \cite{cao-clark-2017-latent} or compute BLEU score among the set of model-generated responses \cite{zhu-et-al-acm}. 
Other work measures the diversity of a \textit{set} of model responses (the \textit{Multi-Response Diversity} setting) \cite{zhang-etal-2019-syntax-infused, tevet-berant-2021-evaluating}.  This setting measures the model's ability to generate multiple diverse responses to a single conversation.

We utilize two automatic diversity metrics in the Multi-Response setting.  
The first metric calculates a diversity score based on a Natural Language Inference (NLI) model's predictions among pairs of utterances 
\cite{nli-placeholder}.  A contradiction prediction is given weight of 1, neutral is 0, and entailment is -1.  The set of scores for all pairwise combinations are summed to produce the final NLI Diversity score. 
The second metric uses Sent-BERT \cite{tevet-berant-2021-evaluating}, based on the similarity of sentence-level BERT embeddings \cite{reimers-gurevych-2019-sentence}.  We calculate  the average pairwise cosine similarity between responses representations, subtracted from 1 to convert the metric from similarity to diversity.

\subsection{Crowdworker Diversity Evaluation}
Past work has used crowdworkers to indirectly measure diversity of model responses, by having workers rate specificity \cite{li-etal-2016-diversity} or interestingness \cite{see-etal-2019-makes} of responses. 
\citet{tevet-berant-2021-evaluating} explore whether crowdworkers can  directly judge diversity of a set of model responses, finding that crowdworkers are able to do this reliably for semantic diversity.
 We draw on this work to explore whether crowdworkers are able to determine the \edivshortspace of input conversations, rather than the diversity of a set of model responses.  In contrast to this work, we rely on an expert crowdworking population of creative writers.

\subsection{Conversation Analysis and Speech Acts}

We  draw from the fields of conversational analysis \cite{schegloff} and pragmatics \cite{levinson_1983}, with a focus on speech acts.  In particular, we consider the notion of \textit{adjacency pairs} (e.g. question-answer, greeting-greeting, and apology-minimization): the combination of two sequential speech acts produced by different speakers, in which the first speech act requires the production of the second \cite{levinson_1983}.  

Throughout this work, we make use of the DailyDialog dataset, which includes manually-labeled high-level speech acts, consisting of one of Inform, Question, Directive, or Commissive \cite{li-etal-2017-dailydialog} (see Table~\ref{tab:speechact_table} for examples). 
Because the DailyDialog speech acts are general, we also make use of classifier-assigned fine-grained speech act categories from the
Switchboard Dialogue Act Corpus \cite{Jurafsky-etal:1997}, which are an extension of the DAMSL coding scheme \cite{core1997} (see Table~\ref{tab:speechact_table} for examples).

\subsection{Evaluating Creativity}\label{sec:creativity}
For our human study, we draw from related work in the measurement of creativity. 
In particular, we use the Creative Assessment Technique (CAT), which establishes a framework to assess creative output \cite{amabile}.  
This process engages experts from  creative fields to independently judge the creativity of different output (e.g., stories, artwork). Usually this framework is deployed to evaluate creative output produced for the same prompt.  However, \citet{baer} showed that CAT produces  consistent judgments for 
products generated in response to different prompts.  

This line of research inspires our work in two ways.  First, we use  expert creative writers instead of typical crowdworkers because 
\citet{kaufman} found that nonexperts produce inconsistent creative judgments.  
We also treat each dialogue as a creative \textit{prompt} and ask writers to rate the extent the prompt inspires the creation of diverse responses.  This task is suggested as future work  in \citet{baer}, but to the best of our knowledge, we are the first to apply it in this manner.

\section{Speech Act Analysis}\label{sec:coarse}
In order to explore whether diversity of sets of responses changes based on the input conversation, we choose the DailyDialog++ dataset, which has two important properties: (i) multiple  human responses for each conversation, and (ii)  high-level speech acts manually assigned to each utterance.  In this section, we  analyze this data using diversity metrics to test hypotheses about \edivshort.

\subsection{Dataset}
The  DailyDialog dataset was scraped from online English learner conversations  and has manually-assigned speech act labels \cite{li-etal-2017-dailydialog}.  DailyDialog++\footnote{\url{https://iitmnlp.github.io/DailyDialog-plusplus/}} augments the original conversations with multiple responses from human annotators.  
 To utilize speech labels, we work with DailyDialog++ data which exactly aligns with the original DailyDialog conversations (9,192 conversations out of the 9,258 in the DailyDialog++ training partition), each with 5 human-generated next responses.

\subsection{Analysis: Human-Labeled Speech Acts}
\begin{table}[]
    \centering
    \small
    \begin{tabular}{|p{1.9cm}|p{1.5cm}|p{1.7cm}|p{0.9cm}|}
      \hline
      \rule{0pt}{1em}\textbf{Speech Act}  &  \textbf{NLI Diversity} 	$\uparrow$ & \textbf{Sent-BERT ~~~~~~~Diversity} 	$\uparrow$ & \textbf{Num. Convos.} \\\hline
        Inform & 4.43 & 0.72 & 3875 \\ 
        Question & 5.72 & 0.71 & 2724\\
        Directive & 5.20 & 0.72 & 1845\\
        Commissive & 4.29 & 0.71 & 748 \\\hline
    \end{tabular}
    \caption{Diversity for multi-response sets responding to different speech acts along with the number of conversations.   
    For both metrics, higher values indicate higher level of diversity.}
    \label{tab:coarse_div}
\end{table}

We hypothesize that the most recent speech act utterance in a conversation will influence the diversity of the set of 5  responses.
We begin by exploring the original DailyDialog labeled speech acts, which fall into one of four categories based on the structure proposed in \citet{amanova-etal-2016-creating}:  Inform, Question, Directive, and Commissive \cite{li-etal-2017-dailydialog}.  Table \ref{tab:speechact_table} contains examples of each speech act with corresponding definitions. 

DailyDialog speech acts are manually assigned to each turn in the conversation; however, for this work we are most interested in the most recent turn.
An assumption that we make throughout this work is that there is one speech act for each utterance, although utterances can in fact have multiple associated speech acts \cite{levinson_1983}.

\subsubsection{Hypotheses}

We explore whether diversity metrics detect how the 5 human-generated responses change based on the most recent speech act utterance.
We hypothesize that Question and Inform will produce more diverse sets of responses than Directive and Commissive, which center around an action which likely constrains the set of responses.  Although Question is part of an adjacency pair, we hypothesize the content of the question can allow for more diverse responses than Directive and Commissive. 

\subsubsection{Results}
To analyze our hypotheses, we examine the diversity distributions for the five multi-response sets for each of the four speech acts.  
For each response set, we calculate a diversity score using either NLI Diversity or Sent-BERT Diversity.  We report the average  diversity scores using both metrics for each speech act in Table \ref{tab:coarse_div}.

While the Sent-BERT diversity scores are similar for all categories, we find larger differences in NLI Diversity between the speech act categories.  In particular, Question is the most diverse, as expected.   Commissive is the least diverse for both metrics, supporting our hypothesis that it would yield less diversity than Question and Inform. 

To test statistical significance, we run a Kruskal-Wallis test \cite{kruskal-wallis} with a Dunn posthoc \cite{dunn} and a Bonferroni adjustment to account for multiple pairwise comparisons \cite{Haynes2013}.
For NLI Diversity, we note significant (p$<$0.05) differences among all pairwise categories except for Inform + Commissive.  For Sent-BERT Diversity, we note significant (p$<$0.05) differences only among Inform + Commissive, Question + Commissive, and Directive + Commissive.

\subsection{Analysis: Model-Labeled Speech Acts}\label{sec:fine}
The speech act categories for DailyDialog++ are human-labeled but quite broad.  Since we are interested in investigating finer-grained categories, e.g. different types of questions,  we also consider a speech act set containing 42 categories:  Switchboard SWBD-DAMSL \cite{Jurafsky-etal:1997, core1997}. While there are domain differences between DailyDailog++ and Switchboard, both  deal with social conversations and we believe have similar speech acts represented.  Additionally, a manual examination of 100 classifications for our speech act subset yields 85 correct predictions.
\begin{figure}[ht!]
    \begin{tabular}{c}
    \includegraphics[width=0.48\textwidth]{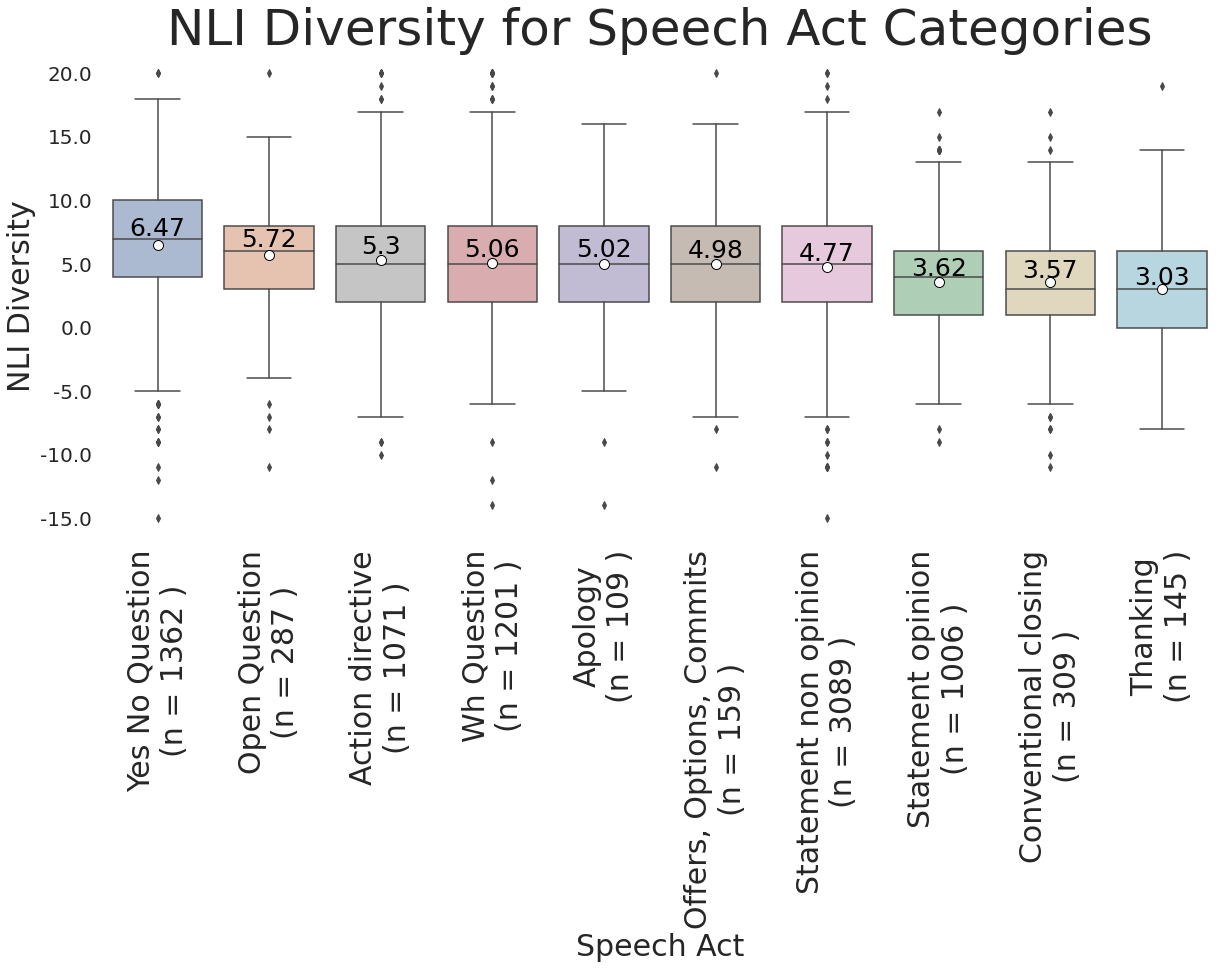} \\
     \\
    \includegraphics[width=0.48\textwidth]{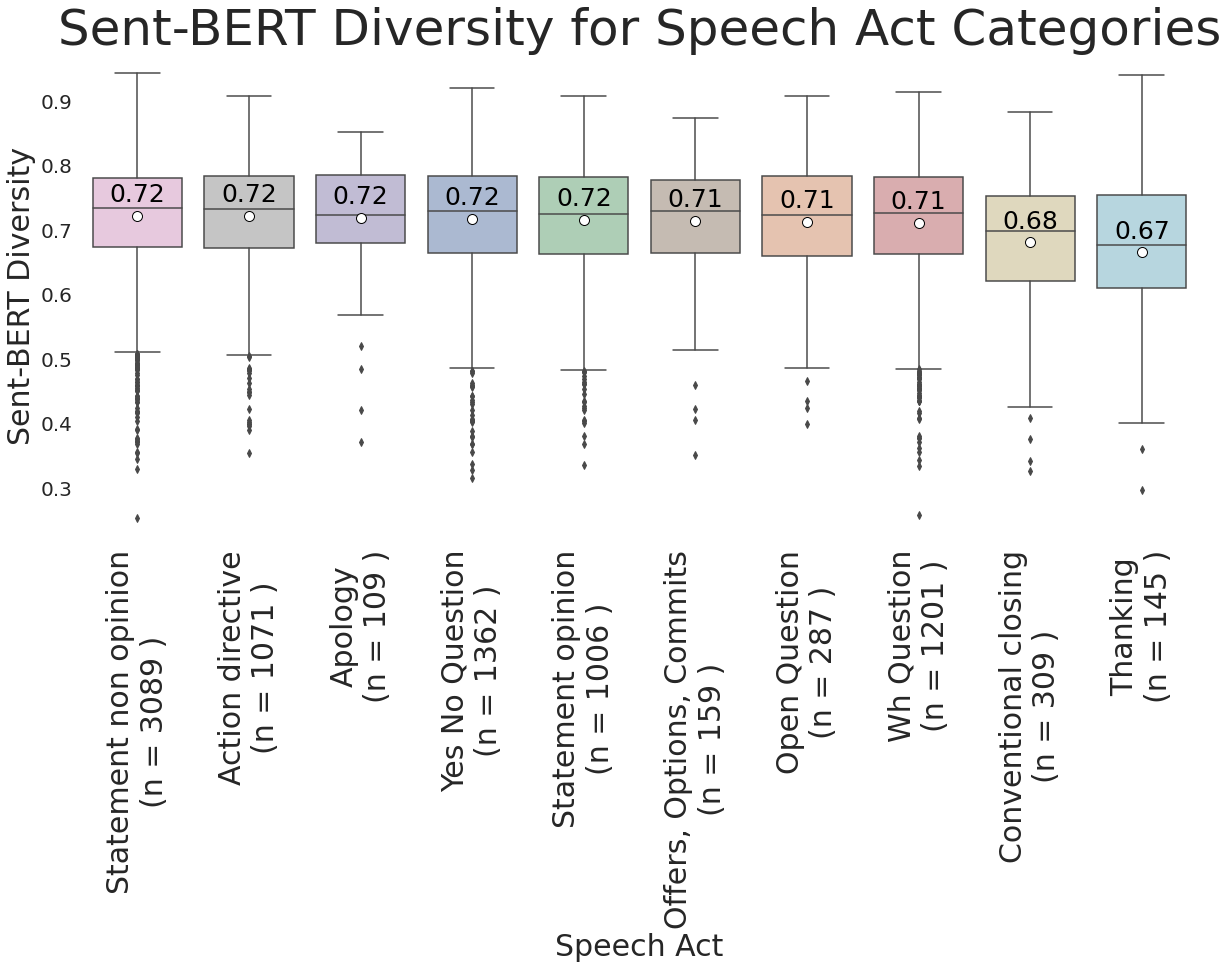}\\
    \end{tabular}
    \caption{NLI Diversity (top) and Sent-BERT (bottom) for responses categorized by most-recent speech act utterance (higher values indicate more diverse, ordered by diversity).  Mean values  are indicated by the white circle and corresponding text label.  Box-and-whisker plots show the interquartile ranges.}
    \label{fig:boxplot_nli}
\end{figure}
\subsubsection{Classification Model}
In order to apply the SWBD classes to the DailyDialog++ dataset, we rely on a speech act classification model.  We use the baseline RoBERTa-based  classification model (125M parameters) from \citet{he-etal-2021-speaker-turn}\footnote{\url{https://github.com/zihaohe123/speak-turn-emb-dialog-act-clf}}, training for 9 hours on a Titan RTX GPU.  Our trained model achieves accuracy of 0.82 and 0.83 when evaluated with gold labels of the SWBD test and validation sets, respectively.  

For each conversation in DailyDialog++, we pass the most recent conversational turn to this classifier, obtaining a speech act classification of one of the 42 categories in SWBD.  We restrict our analysis to 10 acts which are most prevalent in DailyDialog++ (predicted at least 100 times by the classifier across the training set).
We  exclude the ``Continued'' speech act, because it is specific to speech acts continued from a past turn in SWBD and is not applicable to the DailyDialog++ dataset. 

\subsubsection{Hypotheses}
We hypothesize that more constraining speech acts, such as apology, thanking, and conversational closing, will have lower diversity.  Closing-closing, thanking-acceptance, and apology-minimization are all adjacency pairs with constraining responses \cite{levinson_1983}.   Although question-answer is also an adjacency pair, we expect the answer content to contain more diverse content. 
Statements of Opinion and Non-Opinion do not generally constrain responses, so we expect diversity
to vary.

\subsubsection{Results}\label{sec:swbdresults}

The NLI Diversity of DailyDialog++ responses for SWBD categories can be seen in Figure \ref{fig:boxplot_nli} (top).  The highest diversity is seen in response to Yes-No and Open Questions, as expected.  Additionally, as expected, the Thanking and Closing categories result in the lowest NLI Diversity.  Unexpectedly, Apology has a higher diversity than both Statement of Opinion and Non Opinion
; we explore this finding in Section \ref{sec:writers}.  It is also surprising that Yes-No Question falls much higher than Wh Question, likely due to the emphasis on contradictions of the NLI Diversity metric.  In Appendix \ref{app:swbd_sig} we report significance results, finding more significant differences among speech acts using NLI Diversity compared to Sent-BERT.

\section{Study  with Creative Writers}\label{sec:writers}

To explore whether expert judgments of diversity corresponded to the significant NLI Diversity differences observed in Section \ref{sec:fine}, we designed a study to measure \edivshortspace 
from the prior conversation using a novel human evaluation task.

\subsection{Study Design}

\subsubsection{Participants}
Because the task involves judging the creativity of input conversational prompts, we followed recommendations of \citet{kaufman} and  work with expert writers instead of a typical crowdworking population.  To this end, we employed  participants  who have experience with creative writing, screenwriting, or playwriting from the Upwork platform, prioritizing those who had past Upwork  or professional writing experience.  Across all experiments, we worked with a total of 28 writers (two participants repeated one task each on different data).  
Participants were compensated at a rate of \$20 per survey, estimated to take one hour or less.

\subsubsection{Task Design}
Participants completed five sections in the order:  Writing, Drag-and-Drop, Likert, Drag-and-Drop, Likert.
The conditions for each task are:

\textbf{Writing Task:} Participants are instructed to generate ``unique, interesting, and appropriate responses to each dialogue conversation.'' 
Participants generate 5 responses each for four conversations (one from each of the four speech acts in the set), presented in a random order.

\textbf{Drag-and-Drop Task:} 
Participants are  presented with four randomly-ordered conversations (one from each speech act) and asked to drag and drop them such that the top conversation ``most inspires the creation of multiple distinct responses''  and the bottom conversation ``least inspires this.''  

\textbf{Likert Rating Task:}
While the drag-and-drop requires participants to rank conversations against one another, we are also interested in their assessments of conversations in isolation.  Thus, in this task we ask participants to rate 20 conversations on 
a 5-point Likert scale, where 1 represents ``Does not Inspire Creative Responses'' and 5 represents ``Does Inspire Creative Responses.''  We randomize the presentation of the conversations.

\subsubsection{Stimuli Creation}
To examine whether the human writings and judgments can uncover differences between different speech acts, we chose two sets of SWBD speech acts
with varying levels of NLI Diversity from Section \ref{sec:swbdresults}.  
\textit{Set 1} consists of:  Yes-No Question, Wh Question, Thanking, and Apology.  We hypothesize that the Question acts will result in higher \edivshortspace ratings while Thanking and Apology will result in lower ratings, based on trends observed in Section \ref{sec:swbdresults}.  
This set is chosen to display the largest predicted differences between speech acts.

\textit{Set 2} consists of:  Open Question, Opinion Statement, Non-Opinion Statement, and Closing.  We hypothesize that Open Question and Opinion will have higher diversity ratings and Non-Opinion and Closing will have lower ratings.  However, these speech acts were not the most or least diverse.
Therefore, we expect these rating differences to be less pronounced than Set 1's.

We intentionally selected conversations from DailyDialog++ whose NLI Diversity scores
fell into the median range for the speech act, to 
ensure the data collected was from prototypical speech act conversations, not  outliers.  In cases for which sampling from only the median would not result in enough conversations for all surveys, we increased this window to include +/- 3 around the median NLI Diversity value. 
We manually verified the most-recent  turn was classified in the correct speech act category, removing misclassifications.

\subsubsection{Study Conditions}

Each participant completed 52 tasks on different conversations,  evenly distributed among 4 speech acts. 
We constructed 6 Qualtrics surveys which collectively covered 156 conversations (39 per speech act).
We recruited five participants for each survey, resulting in 30 completed surveys.

\subsection{ Study Results}\label{sec:writer_results}
This section presents results for the 
creative writer study.
For the writing task, we find differences that align with our hypotheses; however, significance results are limited.
For both the Drag-and-Drop comparison tasks and the Likert ratings,  we find significant differences for Set 1,  which align with our hypotheses.  Results for Set 2 generally support the hypotheses, with some caveats.  

\subsubsection{Writing Task}

\begin{table}[]
    \centering
    \small
    \begin{tabular}{|p{0.4cm}p{2.4cm}|p{1.4cm}|p{1.8cm}|}
    \hline
    \rule{0pt}{1em}&\textbf{Speech Act} & \textbf{NLI Diversity} $\uparrow$ & \textbf{Sent-BERT Diversity} $\uparrow$\\\hline
   \multirow{4}{*}{\rotatebox[origin=c]{90}{Set 1}}& Yes-No Question & 7.3 & 0.74 \\
    & Wh Question & 4.8 & 0.73 \\
    & Apology & 2.3 & 0.65 \\
    & Thanking & 2.6 & 0.66 \\\hline
     \multirow{4}{*}{\rotatebox[origin=c]{90}{Set 2}}& Open Question & 6.2 & 0.77 \\
    & Opinion & 3.9 & 0.74 \\
    & Non-Opinion & 6.0 & 0.74 \\
    & Closing & 4.7 & 0.74\\\hline
    \end{tabular}
    \caption{Average NLI and Sent-BERT Diversity among sets of participant-produced responses.  Speech acts are ordered within each set based on our hypothesized diversity level, from most to least diverse.}
    \label{tab:writing}
\end{table}

We begin by measuring the diversity of the sets of 5 responses produced in the Writing component of the survey. 
Table \ref{tab:writing} reports the average NLI and Sent-BERT diversity of these produced responses; Appendix \ref{app:writing} contains responses. Our analysis with responses produced by creative writers replicates the results we saw with the DailyDialog++ responses: 
For Set 1, Yes-No and Wh are 
more diverse than Apology and Thanking. 
For Set 2, Open Question is more diverse  than Closing, also as expected.  Unexpectedly, Non-Opinion and Closing are more diverse than Opinion using NLI Diversity, but the difference is relatively small.  

For the NLI Diversity of the Set 1 responses, we find significant results (p $<$ 0.05) using Kruskal-Wallis with a Dunn posthoc with Bonferroni adjustment between Yes-No Question + Apology.  
Although a Kruskal-Wallis test yields p$<$0.05 for Set 1 + Sent-BERT, a Dunn posthoc with Bonferroni adjustment does not yield significant pairwise results.
For  Set 2, for both metrics, the initial Kruskal-Wallis test  did not show significance so we did not compute pairwise results.
A possible explanation for the lack of significance 
for Sent-BERT is that we selected 
data based on the median \textit{NLI Diversity} values of response sets; however, future work could replicate this experiment with median Sent-BERT conversations to see if significance emerges.

\subsubsection{Drag-and-Drop Task}\label{sec:draganddrop}

\begin{figure*}[htp!]

\begin{subfigure}{\textwidth}
\includegraphics[width=\textwidth]{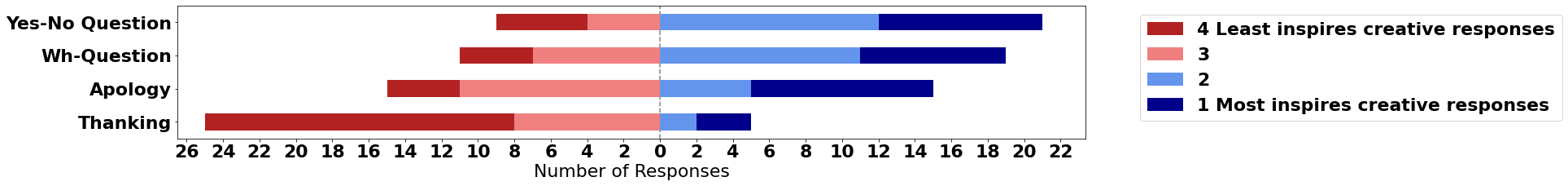}
\end{subfigure}

\medskip

\begin{subfigure}{\textwidth}
\includegraphics[width=\textwidth]{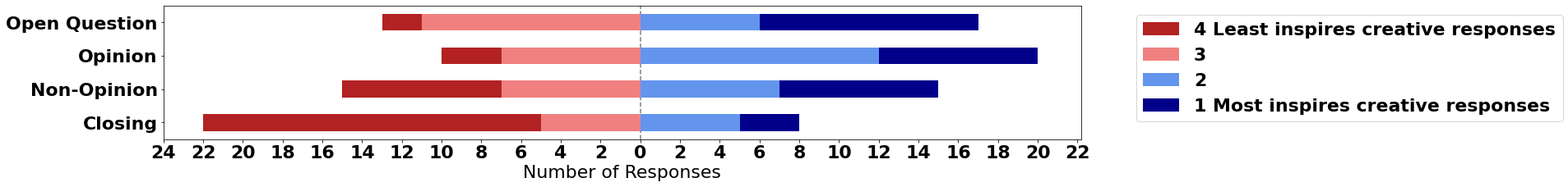}
\end{subfigure}

\caption{Drag-and-drop results for Set 1 (above) and Set 2 (below). } 
\label{fig:rank}
\end{figure*}

Results from the drag-and-drop section can be seen in Figure \ref{fig:rank}.  For Set 1, the highest-ranked speech act is Yes-No Question, followed by Wh-Question, Apology, and Thanking, which confirms our observed NLI Diversity for each speech act from Section \ref{sec:swbdresults}.  A Friedman's significance test \cite{friedman} with Nemenyi posthoc \cite{nemenyi1963distribution} finds that Yes-No Question, Wh-Question, and Apology are rated significantly higher than Thaking, confirming our hypothesis.

For Set 2, we observe the Closing speech act results in the lowest ratings.  Surprisingly, the Opinion conversations were rated more diverse than the Open Question conversations.  Using a Friedman's significance test with Nemenyi posthoc, however, we only find significant differences between Open Question + Closing and Opinion + Closing.  This aligns with our hypothesis that Closing conversations constrain the diversity of responses.

\begin{figure*}[htp]

\begin{subfigure}{\textwidth}
\includegraphics[width=\textwidth]{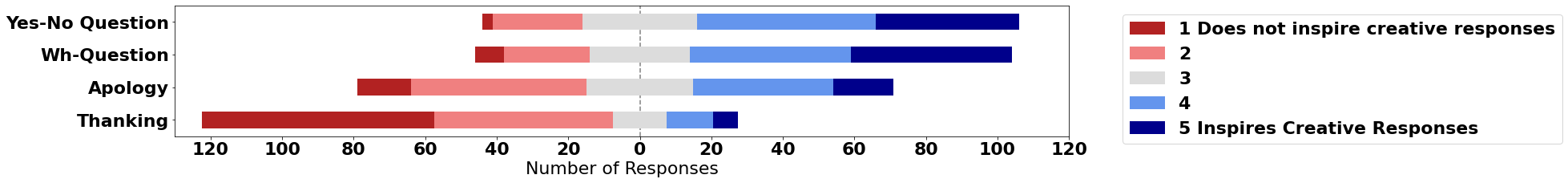}
\end{subfigure}

\medskip

\begin{subfigure}{\textwidth}
\includegraphics[width=\textwidth]{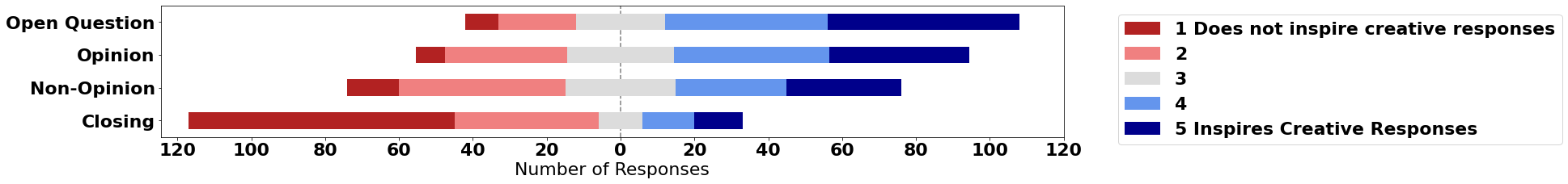}
\end{subfigure}

\caption{Likert results for Set 1 (above) and Set 2 (below).}
\label{fig:likert}
\end{figure*}

\subsubsection{Likert Rating Task}
Likert ratings for each set of speech acts are shown in Figure \ref{fig:likert}; Appendix \ref{app:likert} contains scored conversations.  For Set 1, we note highest ratings among Yes-No and Wh-Questions, lowest ratings with Thanking, and middle ratings with Apology.  This is similar to the drag-and-drop results (Section \ref{sec:draganddrop}). A Friedman's significance test with Nemenyi posthoc yields significance (p $<$ 0.05) among all pairwise speech acts except for Yes-No + Wh Question.

For Set 2, ratings also aligned with our hypotheses, with Open Question, Opinion, Non-Opinion, and Closing ordered from most- to least-diverse.  Contrary to the drag-and-drop results in Section \ref{sec:draganddrop}, we find Open Questions are rated higher than Opinion conversations in the Likert section.  We find significant differences (p < $0.05$) between all pairwise speech act combinations except for Open Question + Opinion and Opinion + Non-Opinion.
In Appendix \ref{app:likert_metric} we explore the correlation between Likert ratings and diversity metrics, finding a moderately positive correlation with NLI Diversity and a very weak positive correlation with Sent-BERT.

\section{Discussion and Future Work}\label{sec:discussion}

We found that controlling for speech act type resulted in significant differences in \edivshortspace using automatic diversity metrics (Section \ref{sec:swbdresults}) as well as   creative writing participant judgments (Section \ref{sec:writer_results}).  In particular, certain types of   speech act utterances (Thanking, Apology, and Closing) were consistently judged less likely to produce diverse responses than other speech act utterances. 

\textbf{Implications.}
One important implication of this work is the finding that diversity is sensitive to the pragmatics of the utterance being responded to.
This paves the way for pragmatics to be incorporated into model evaluation, e.g. by defining the \edivshortspace for  conversations in the evaluation dataset and then assessing whether a model's diversity correlates with  \edivshortspace scores.

These results  have implications for  dialogue model generation in addition to evaluation.  Perhaps a simpler, rule-based system should be used for pragmatically-constraining conversations, relying on a neural network generation system only for conversations with high \edivshort.  It is important to note that conversational creativity varies; one participant in the study rated every conversation with a score of 5, and wrote:
``I tend to think creative options are without limit, especially with limited context.''  When creating and assessing dialogue systems, we should consciously choose if we are creating a witty Oscar Wilde or a conversational partner who follows pragmatic norms.

Another important finding of this work is that overall, the NLI Diversity metric, which explicitly incorporates contradiction and entailment predictions, was more distinguishing than the Sent-BERT measure, which 
supports results found in \citet{nli-placeholder}.  

\textbf{Future Work.}  Our  findings examined eight different speech acts in depth, but  additional common speech acts can be explored, such as  Action Directives. 
While we found significant differences among diversity relating to the most recent speech act utterance, it is possible that the other speech acts in the conversation influence diversity. Future work should also  examine the effects of other variables, such as length or the topic of conversation.

To circumvent the need for expert labeling of speech acts, future work should investigate automatically predicting the 5-point \edivshortspace rating from expert data. 
A baseline approach to this task could use the median for the speech act utterance.

\section{Conclusion}\label{sec:conclusion}
We introduce \ediv, the hypothesis that 
dialogue diversity will vary based on the speech act type of the input conversation.  An analysis finds significant differences for automatic diversity metrics among both human-labeled speech acts from DailyDialog++ and fine-grained speech acts automatically assigned by a SWBD model.  We use these findings to create a new human evaluation task, to explore whether the assessments by creative writers align with a conversation's \edivshort.  We find that writer rankings correspond with our hypotheses, paving the way for \edivshortspace 
to be incorporated into future dialogue evaluation and generation.

\section*{Limitations}
Our experiments with speech acts are all conducted with human-generated dialogue, instead of model-generated dialogue.  While this design decision allows us to demonstrate that \edivspace differs even among human responses, it means that our results do not include model-specific generation issues, such as hallucinating information from training data \cite{roller-etal-2021-recipes} or their ability to generate harmful data in response to adversarial attacks \cite{xu-etal-2021-bot}.  However, it is important to verify that \edivshortspace can detect differences with human-generated dialogue data before exploring neural responses; that  can be done in future work. 
Our approach can also be combined with other work in dialogue which aims to address hallucination and harmful data production.

While our work provides evidence that \edivspace differs based on the most recent speech act utterance and that this should potentially be incorporated into diversity evaluation of dialogue models, we do not assess it in a  downstream application, leaving  this to future work.  

Additionally, our work only covers a subset of speech acts found in in SWBD, which can be expanded in future work.  We are also working with one conversational dataset because it had multiple human-generated responses along with speech act labels; additional experimentation should expand this to other datasets.

We conduct experimentation with a conversational English dataset and motivate our work using pragmatic literature grounded in English interactions.  It is possible that the \edivspace of other languages would  differ from English \ediv, especially in terms of the particular speech acts and adjacency pairs given documented differences in other languages \cite{huth}.

\section*{Acknowledgements}
This work was supported by an AWS Machine Learning Research Award, an NVIDIA Corporation GPU grant, an AI2 Key Scientific Challenge Proposal grant, and a National Science Foundation (NSF) Graduate Research Fellowship (DGE 1752814). We thank the 28 writers for their participation as well as Philippe Laban, Nate Weinman, and the Hearst Lab Research Group for their helpful comments.

\bibliography{anthology,custom}
\bibliographystyle{acl_natbib}
\clearpage
\pagebreak
\appendix

\section{Most Recent Speech Act Utterances with Full Conversations}\label{app:speechact_convos}

Table \ref{tab:speechact_full_convos} contains full conversations where the most recent utterance is classified as different speech acts.

\begin{table*}[ht!]
\small
    \centering
    \begin{tabular}{|p{3cm}|p{11.5cm}|}
     \multicolumn{2}{c}{\textbf{DailyDialog++ Speech Act Utterance Example Conversations}} \\\hline
     \rule{0pt}{1em}
     \textbf{Inform} & \textbf{Speaker 1:} I have children so a steady job is important but I would like a chance to advance.\newline 
\hspace*{0.5cm} \textbf{Speaker 2:} How would you feel about relocating to another state?\newline 
\textbf{Speaker 1:} I could not relocate in the next year, but might be open to it in the future.\\\cline{2-2}
     \textbf{Question} & \textbf{Speaker 1:} Do I often have to work overtime?\newline 
\hspace*{0.5cm} \textbf{Speaker 2:} Yes, you have to work overtime a lot due to the editing job.\newline 
\textbf{Speaker 1:} Is there any extra pay for that? \\\cline{2-2}
     \textbf{Directive} & \textbf{Speaker 1:} What's the most popular paper circulating in the community?\newline 
\hspace*{0.5cm} \textbf{Speaker 2:} Atlanta Daily.\newline 
\textbf{Speaker 1:} Oh, thank you . I want one.\\\cline{2-2}
     \textbf{Commissive} & \textbf{Speaker 1:} Don't take any chances . It'd be best if you told him and promised not to be late again . He's already mad at you for last 2 times . Any more and he might fire you.\newline 
\hspace*{0.5cm} \textbf{Speaker 2:} You said it . It won't happen again . Do you really think he'd fire me?\newline 
\textbf{Speaker 1:} I think he might . You'd better go to his office.\newline 
\hspace*{0.5cm} \textbf{Speaker 2:} You are probably right . I'll go right now and apologize . I try hard not to be late but it is difficult with Beijing traffic.\\\hline
     \multicolumn{2}{c}{}\\
     \multicolumn{2}{c}{\textbf{SWBD Speech Act Utterance Example Conversations (Selected Subset)}}\\\hline
      \rule{0pt}{1em}\textbf{Statement Non Opinion}   & \textbf{Speaker 1:} Where are you from, Laura?\newline 
\hspace*{0.5cm} \textbf{Speaker 2:} Well, my whole family is in the United States now, but we're from Costa Rica originally.\newline 
\textbf{Speaker 1:} Oh, so you're from South America.\newline 
\hspace*{0.5cm} \textbf{Speaker 2:} Actually, Costa Rica isn't in South America . It's in Central America. \\\hline
     \textbf{Yes-No Question}   & \textbf{Speaker 1:} Excuse me . Is anyone in that seat next to you?\newline 
\hspace*{0.5cm} \textbf{Speaker 2:} No, I don't think so.\newline 
\textbf{Speaker 1:} Do you mind if I put my jacket there? \\\hline
     \textbf{Wh Question}    & \textbf{Speaker 1:} Let me ask some questions . Do you know what people need to get entry into China?\newline 
\hspace*{0.5cm} \textbf{Speaker 2:} Yes . An passport and a visa are necessary for entry into China.\newline 
\textbf{Speaker 1:} What materials do they need in order to apply the passport? \\\hline
     \textbf{Action Directive}   & \textbf{Speaker 1:} Would you mind showing me your passport?\newline 
\hspace*{0.5cm} \textbf{Speaker 2:} Here it is . And how much RIB shall I get?\newline 
\textbf{Speaker 1:} Let me see . \$200 makes RIB 1,616 . Here is the cash and exchange memo . Please check it. \\\hline
     \textbf{Statement Opinion}    & \textbf{Speaker 1:} Alice, I never knew you had such a lovely voice . You really can sing, can't you?\newline 
\hspace*{0.5cm} \textbf{Speaker 2:} Thanks, Mark . I used to be a member of the school choir.\newline 
\textbf{Speaker 1:} No wonder you can control your voice so well . You are a professional singer.\newline 
\hspace*{0.5cm} \textbf{Speaker 2:} Well, you are flattering me . I wouldn't say I am a professional, but I did receive some training at school . My music teacher used to be a professional singer.\newline 
\textbf{Speaker 1:} Well, a good teacher makes good students.\\\hline
     \textbf{Conventional Closing}    & \textbf{Speaker 1:} Well, I must be off now . I have an appointment at six.\newline 
\hspace*{0.5cm} \textbf{Speaker 2:} In that case, I won't keep you any longer . Drop in any time.\newline 
\textbf{Speaker 1:} Ok . Goodbye. \\\hline
     \textbf{Open Question}   & \textbf{Speaker 1:} Are you have a hand in locking into the case?\newline 
\hspace*{0.5cm} \textbf{Speaker 2:} Yes.\newline 
\textbf{Speaker 1:} How do you feel about it? \\\hline
     \textbf{Offers, Options, Commits}    & \textbf{Speaker 1:} I can understand your position . Perhaps I'm asking too much.\newline 
\hspace*{0.5cm} \textbf{Speaker 2:} We'll use wooden cases if you insist, but the charge for packing will be considerably higher, and it also slows delivery.\newline 
\textbf{Speaker 1:} Well, I'll call you immediately for instructions on the matter. \\\hline
     \textbf{Thanking}    & \textbf{Speaker 1:} Oh! That's much hetter than I did.\newline 
    \hspace*{0.5cm} \textbf{Speaker 2:} Can you tell me how to improve myself?\newline 
\textbf{Speaker 1:} There's no secret at all . The only thing for you to do is to practice more . You will succeed.\newline 
\hspace*{0.5cm}\textbf{Speaker 2:} Thank you for your encouragement. \\\hline
     \textbf{Apology}    & \textbf{Speaker 1:} So, Is this your first time to TAIWAN?\newline 
\hspace*{0.5cm} \textbf{Speaker 2:} No, I first came here 1995.\newline 
\textbf{Speaker 1:} Oh, really? And you are from the State, right?\newline 
\hspace*{0.5cm} \textbf{Speaker 2:} Well, I am from Canada . Actually.\newline 
\textbf{Speaker 1:} Oh, I am so sorry.\\\hline
    \end{tabular}
    \caption{Full conversations for most recent speech act utterances.}
    \label{tab:speechact_full_convos}
\end{table*}

\section{SWBD Significance Analysis}\label{app:swbd_sig}
\begin{figure}[h!]
    \begin{tabular}{c}
    \includegraphics[width=.45\textwidth]{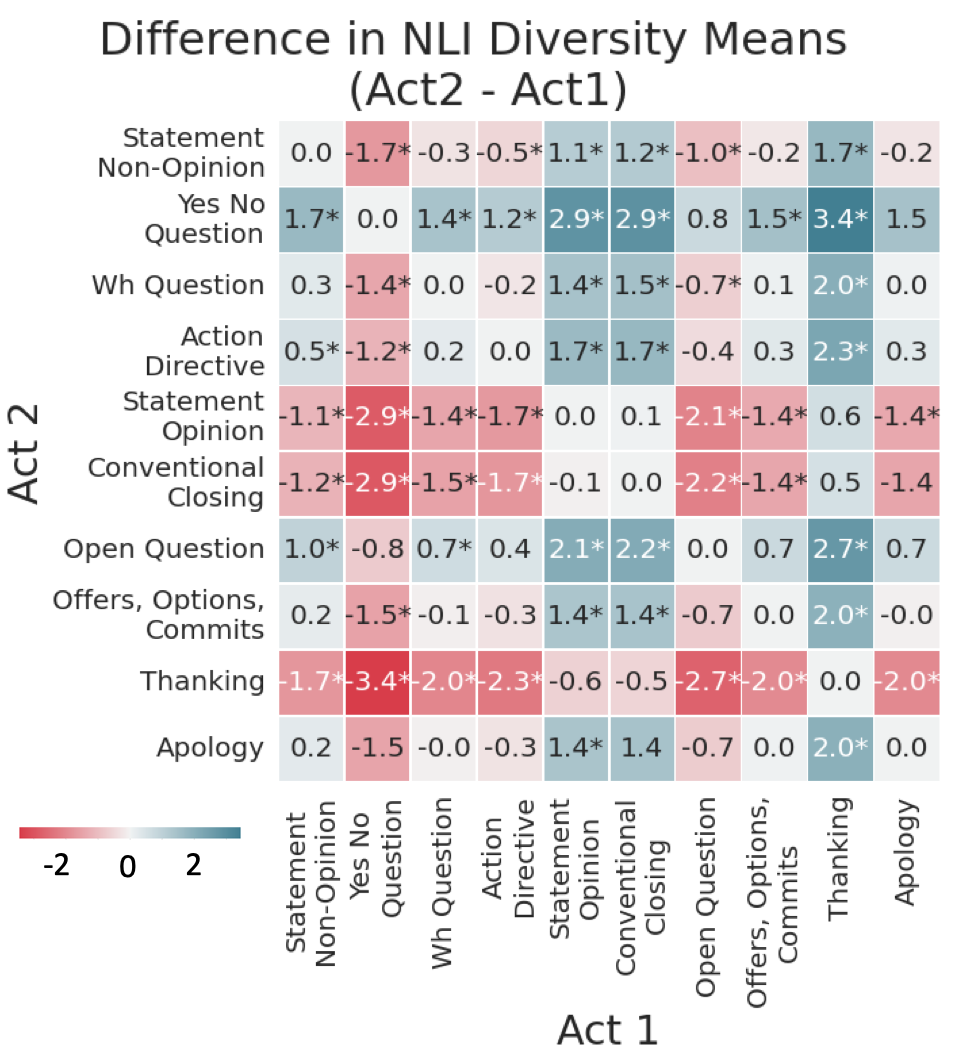}
    \\
    \\
    \includegraphics[width=0.45\textwidth]{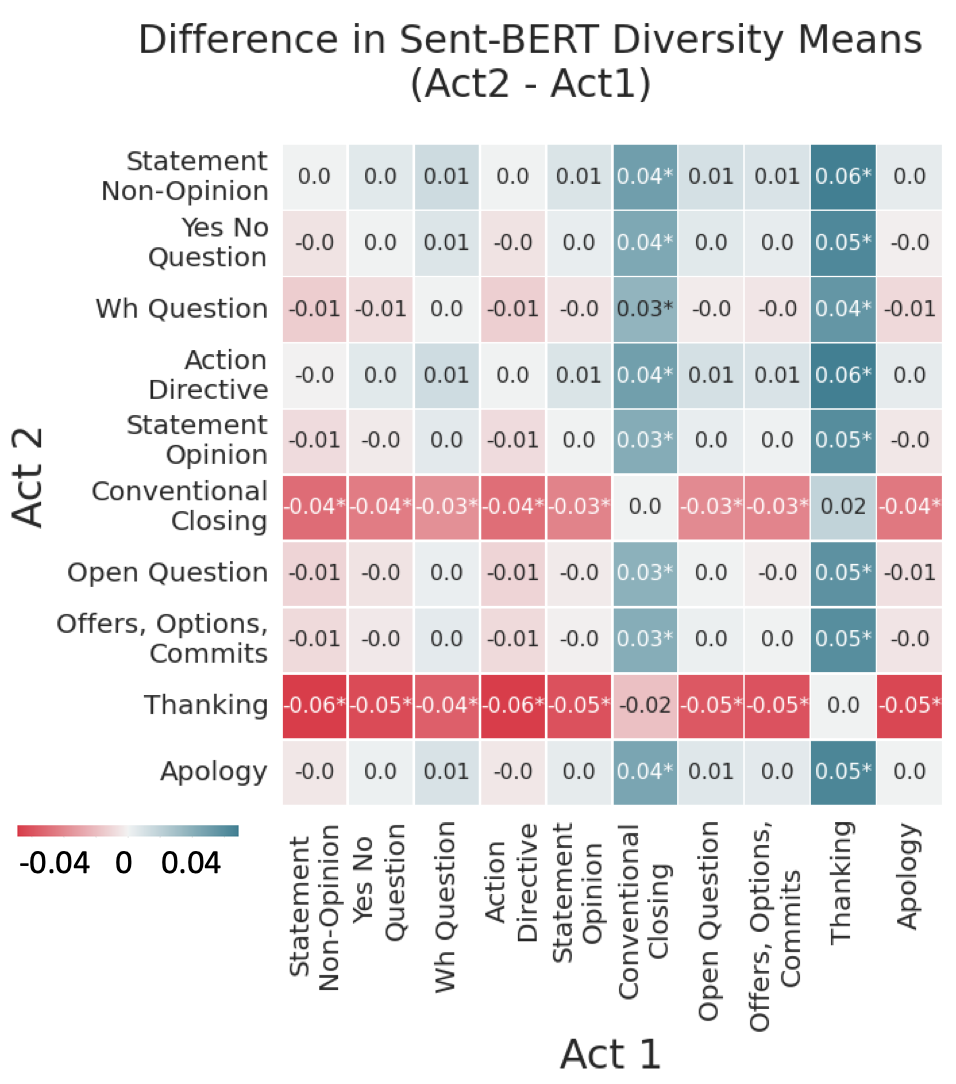}\\
      \end{tabular}
    \caption{(top) Differences in NLI Diversity means between different speech acts. (bottom) Differences for Sent-BERT Diversity means. For both plots, * indicates significance with p$<$0.05.  Darker blue rows indicate the Act2 speech act is more diverse than the Act1 speech act.}
  
    \label{fig:nli_sig}
\end{figure}

We report pairwise difference of means with corresponding significance levels for NLI Diversiy in Figure \ref{fig:nli_sig} (top).  Looking at a row of this figure, the presence of blue (positive) values indicates the speech act is more diverse; in contrast, the presence of red (negative) values indicates the speech act is less diverse.  For example, the ``Yes No Question'' row 
has high diversity and the ``Thanking'' row 
has low diversity.

Significance for Figure \ref{fig:nli_sig} is computed with a Kruskal-Wallis test, Dunn posthoc, and Bonferroni adjustment, which accounts for  the amount of pairwise comparisons being made among speech act categories.  Significant results have p $<$ 0.05.  We note Yes-No Question, Wh-Question, Action Directive, and Open-Question diversity is significantly higher than most other categories while Thanking and Closing are significantly lower.  These significance results support our hypotheses about different diversity scores in response to different speech acts.

We additionally compare Sent-BERT diversity scores to NLI scores to see if the differences in diversity for different speech acts are consistent across multiple diversity metrics.  Sent-BERT results for SWBD categories can be seen in Figure \ref{fig:boxplot_nli} (bottom).  Sent-BERT scores are more similar across categories, with the exception of Closing and Thanking which are lower.  

To explore significance among categories for the Sent-BERT comparison, we perform a Kruskal-Wallis test with Dunn posthoc and Bonferroni adjustment.  We present mean differences with significance results in Figure \ref{fig:nli_sig} (bottom).  We note the low (red) rows are Conventional Closing and Thanking, similar to NLI Diversity findings.  
Overall, we find that Sent-BERT diversity scores give less insight into potential differences between speech act classes than NLI Diversity.  Thus, for the next section, we chose to use NLI Diversity to explore the creative writing task.

\section{Writing Task Data}\label{app:writing}
Table \ref{tab:writing_task_data} contains conversations with corresponding writer responses for each Set of speech acts. The writer responses are reported in the same order within each Set so the writer responses to different speech acts can be compared.

\clearpage
\onecolumn
 \small

\begin{longtable}{|p{2cm}|p{3cm}|p{10cm}|}
 \hline\textbf{Speech Act} & \textbf{Past Conversation} & \textbf{Writer-Generated Responses}\\\hline
 Yes-No Question &  \textbf{Speaker 1:} Honey, where did you buy this skirt? \newline \hspace*{0.5cm} \textbf{Speaker 2:} In the market . Any problem? & $\bullet$ No, no problem. I was just curious. \newline $\bullet$ No, I really like it!\newline $\bullet$ When did you go to the market?	\newline $\bullet$ No, but I thought you were going to the grocery store!	\newline $\bullet$ Not really, but do you need another skirt. \newline \newline $\bullet$ Which market?	\newline $\bullet$ It's stained right here. You should take it back.\newline $\bullet$ You already have one just like it. \newline $\bullet$ It's the same as the one she has been looking for.	\newline $\bullet$ Why not another color? I don't like it.\newline\newline $\bullet$ No, it's just very unique and I had to say something about it!	\newline $\bullet$ Do you think there's a problem with it?\newline $\bullet$ Oh, that makes sense. It has some loose threads. I was wondering if it was a thrift item.\newline $\bullet$ Nope, I was just making small talk. I'm a bit frazzled today and I can't think of much to say.\newline $\bullet$ Yes, it doesn't go well with your complexion. Perhaps you should find something in a different color.\\\hline
 Wh Question & \textbf{Speaker 1:} Do you want a taxi? \newline \hspace*{0.5cm} \textbf{Speaker 2:} Yes. \newline \textbf{Speaker 1:} Where to? &  $\bullet$I need to go to the train station.  \newline $\bullet$ I would like to go to the mall on 3rd Street.  \newline $\bullet$ Can you take me to the airport?  \newline $\bullet$ 19822 63rd Street, please.  \newline $\bullet$ To the museum near the lake. \newline \newline $\bullet$The bowling alley down on 8th street, please. \newline $\bullet$ How far east can you go? I've got cash to cover it. \newline $\bullet$ I don't believe that is any of your concern. \newline $\bullet$ I'll see when he texts me back. \newline $\bullet$The place we went yesterday. What was it called? \newline \newline $\bullet$  I'm visiting the city with a friend but their flight doesn't arrive until tomorrow so I'm sightseeing solo today. What's your favorite area of the city? \newline $\bullet$ I'm headed to work but first I need to stop at the nearest coffee shop. \newline $\bullet$ The nearest nature park, please. I'm in desperate need of fresh air! \newline $\bullet$  I'm not sure. I want to but a few books today but I'm new to the area and wanted to ask a local's opinion on the best used bookstore. I don't want to fund any large corporations. Do you know of a place? \newline $\bullet$ The courthouse. I have a fine to pay. Apparently, it's illegal to leave a couch out on the curb for collection with the garbage. Who knew?!\\\hline
 Apology &  \textbf{Speaker 1:} Please come in and sit down . I ’ m happy to finally meet you. \newline \hspace*{0.5cm} \textbf{Speaker 2:} Same here, Ms . Drake . I've been looking forward to this. \newline \textbf{Speaker 1:} I ’ m sorry I kept you waiting. & $\bullet$ No worries, I was just finishing up some work. \newline $\bullet$ I was running a little late myself! \newline $\bullet$ It's okay! I am glad that we are finally able to meet! \newline $\bullet$ Unfortunately, I have another meeting at 1:00 so I don't have too much time. \newline $\bullet$ I was beginning to wonder if you forgot about our meeting! \newline \newline $\bullet$ It's not a problem, I didn't wait very long. \newline $\bullet$ Just don't keep me waiting again. \newline $\bullet$ He said I should have expected you to be late. \newline $\bullet$ Don't worry, I know how the trains can be this time of day. \newline $\bullet$ Never mind that. Let's not waste any more time.\newline \newline $\bullet$ No worries! It's hectic times these days! \newline $\bullet$ I understand completely. Please, don't feel guilty about rescheduling. \newline $\bullet$ It's no bother. I'm busy too but I always have time for you! \newline $\bullet$Not at all. Now, what would you like to discuss? \newline $\bullet$Admittedly, it wasn't a great first impression, but you can make up for that now. What about ordering some lunch for the two of us?\\\hline
 Thanking & \textbf{Speaker 1:} Would you care to make another appointment? \newline \hspace*{0.5cm} \textbf{Speaker 2:} Unfortunately, I'm leaving town rather unexpectedly, and I may be gone for several days. \newline \textbf{Speaker 1:} I see . I can tell Mr.Thomas that you phoned. \newline \hspace*{0.5cm} \textbf{Speaker 2:} Thank you. & $\bullet$Your welcome. \newline $\bullet$ Is there anything else that I can help with? \newline $\bullet$ What is the number that Mr. Thomas can rea \newline $\bullet$ Please try to call at least a week in advance if you decide to make another appointment. \newline $\bullet$ Would you like to leave a message for Mr. Thomas? \newline \newline $\bullet$ No problem. Have a good rest of your day. \newline $\bullet$Would you like to make an appointment if he's available when you return? \newline $\bullet$ He'll call you to discuss your next appointment. \newline $\bullet$ Is that all for today? \newline $\bullet$ Would you be interested in a phone appointment if he has one? \newline \newline $\bullet$ Is there any other message that you would like me to relay to Mr. Thomas? \newline $\bullet$You are welcome. Will that be all? \newline $\bullet$ Have a nice day, goodbye. \newline $\bullet$ Of course, have a safe trip! \newline $\bullet$ Would you like to get a call back from Mr. Thomas?\\\hline\hline
 Open Question & \textbf{Speaker 1:} How often? \newline \hspace*{0.5cm} \textbf{Speaker 2:} Three times a year at most . How do you feel about this? & $\bullet$ That's fine with me. \newline $\bullet$ That sounds a little excessive. How about two? \newline $\bullet$ Only three times a year!? Sign me up! \newline $\bullet$ I don't think your parents need to come into town that frequently. An annual visit around the holidays should suffice. \newline $\bullet$It really doesn't matter to me. Whatever you think is best. \newline \newline $\bullet$ Three times a year seems sporadic; what's the catch? \newline $\bullet$ How much is the pay again? \newline $\bullet$ I don't know; would I always be on call? \newline $\bullet$ I'd need to know the exact dates. I will have to check my schedule. \newline $\bullet$ I fell pretty great about it!	\newline \newline $\bullet$ That seems like a lot. I don't even want to go once. \newline $\bullet$ THREE??? \newline $\bullet$ Oh, only three? That's not so bad. \newline $\bullet$ I wish it would be more often. \newline $\bullet$How do YOU feel about it? \\\hline
 Opinion & \textbf{Speaker 1:} I'm looking forward to our son's graduation this weekend. \newline \hspace*{0.5cm} \textbf{Speaker 2:} Yes . So am I . But what will he do after graduation? He really needs to go to college. \newline \textbf{Speaker 1:} Well, dear, we can't force him to go to college . It's up to him. &  $\bullet$I understand, but as a parent, I can't help but worry. \newline $\bullet$True, but I think he'll have better career opportunities if he goes. \newline $\bullet$What do you mean? I thought we were on the same page about this. \newline $\bullet$I know. I just want him to do whatever makes him happy. \newline $\bullet$You're right. Honestly, I'm just ready to have the house back to ourselves.	\newline \newline $\bullet$He's your son! You can threaten to cut him off if he refuses to do what you want. \newline $\bullet$ Of couse, you can't force him. But you want the best for him, don't you? \newline $\bullet$Children should respect their elders and listen to their sage advice. \newline $\bullet$Your son is a smart kid; the world deserves to see what he has to offer. \newline $\bullet$What's the alternative? A trade? He at least needs to know what direction his life will be taking. \newline \newline $\bullet$I know, but I'm just worried and want him to have a good future. \newline $\bullet$I know, but I don't want to give him all that money we saved up unless it's for college. \newline $\bullet$I think that girlfriend of his is influencing his decisions. \newline $\bullet$That's true. And he does have a great job here in town if they take him on full time. \newline $\bullet$I don't understand why he doesn't want to go -- we both did! \\\hline
 Non-Opinion & \textbf{Speaker 1:} We scored six goals. \newline \hspace*{0.5cm} \textbf{Speaker 2:} I don't believe it. \newline \textbf{Speaker 1:} And I score three of them. \newline \hspace*{0.5cm} \textbf{Speaker 2:} You are not serious? \newline \textbf{Speaker 1:} We had extra time of course. & $\bullet$And how many points did the other team have? \newline $\bullet$That's a big improvement! I see all the extra practice is paying off. \newline $\bullet$How much extra time did you have? \newline $\bullet$It sounds like it was a great game. I'm sorry I missed it. \newline $\bullet$Finally! You guys have been on a real losing streak. \newline \newline $\bullet$Extra time should be against the game's rules!	You're the best goalie this team has seen; I'm proud of you! \newline $\bullet$Only with extra time would you manage to win. \newline $\bullet$I can't believe you scored that many goals with that injury. \newline $\bullet$That's why the coach should have made you first string last season	\newline \newline $\bullet$Yeah right! \newline $\bullet$How much extra time? \newline $\bullet$How many of your goals were in extra time \newline $\bullet$How close was the game? \newline $\bullet$No way. \\\hline
 Closing & \textbf{Speaker 1:} I just dropped in to say good-bye. \newline \hspace*{0.5cm} \textbf{Speaker 2:} What time are you leaving? \newline \textbf{Speaker 1:} I'm going to try to leave by ten. \newline \hspace*{0.5cm} \textbf{Speaker 2:} Take care and give my best to your parents. \newline \textbf{Speaker 1:} Good-bye . Hope to see you soon again next year. & $\bullet$Ciao! \newline $\bullet$I hope so too. Feel free to give me a call over the break. \newline $\bullet$Goodbye! I'll see you in a few months! \newline $\bullet$Take care of yourself. \newline $\bullet$I'll hit you up on Instagram!	\newline \newline $\bullet$Next year? \newline $\bullet$I was hoping that we could see each other sooner than that... \newline $\bullet$You can't be too busy to spend time with your best friend, can you? \newline $\bullet$Good-bye; I'll see you in another year, I guess. \newline $\bullet$Try to take more time off, will you? We miss you around here.\newline \newline $\bullet$Yeah! I hope so too! \newline $\bullet$I hope so too, drive safe! \newline $\bullet$Good-bye. Will you let me know when you get back? \newline $\bullet$Okay, we'll see you next year! \newline $\bullet$Hopefully sooner than that! \\\hline
  \caption{Selected conversations with three crowdworker responses for each Task.  Within each Task, the order of response sets is consistent among crowdworkers so responses to different speech acts can be compared.}
 \label{tab:writing_task_data}
 \end{longtable}
\clearpage
\normalsize
\twocolumn
\section{Likert Rating Task Data}\label{app:likert}
Table \ref{tab:likert_full_convos} contains conversations from our tasks with the corresponding Likert ratings assigned by creative writers.

\begin{table*}
\center
\small
\begin{tabular}{|p{2.5cm}|p{7cm}|p{2.5cm}|}
\hline \textbf{Speech Act} & \textbf{Conversation} & \textbf{Ratings} \\\hline
Yes-No Question & \textbf{Speaker 1:} Is that how you feel on the bus? \newline \hspace*{0.5cm} \textbf{Speaker 2:} Uh, well, no, not really, because I haven't worked in England for a long time, so I haven't taken a bus for a long time. \newline \textbf{Speaker 1:} So, so you're British, do you missing them? & 3, 4, 4, 5, 5 \\\hline
Wh-Question & \textbf{Speaker 1:} Ant Shirley, it's being years since we last met . How were you doing in the passing years? \newline \hspace*{0.5cm} \textbf{Speaker 2:} Pretty well . What about you? \newline \textbf{Speaker 1:} Fine . Where are the other guys? & 3, 4, 5, 5, 5 \\\hline
Apology & \textbf{Speaker 1:} Can I make preservation now? \newline \hspace*{0.5cm} \textbf{Speaker 2:} Sure . At what time? \newline \textbf{Speaker 1:} 9:45, please . We want a table next to the window. \newline \hspace*{0.5cm} \textbf{Speaker 2:} I am sorry, sir . We are booked until 9:10. & 2, 3, 3, 3, 4 \\\hline
Thanking & \textbf{Speaker 1:} Sounds good. \newline \hspace*{0.5cm} \textbf{Speaker 2:} Thanks for your time. & 1, 1, 1, 1, 5 \\\hline\hline
Open Question & \textbf{Speaker 1:} Don't worry, sir . I've never cut a customer . Shall I trim your moustache? \newline \hspace*{0.5cm} \textbf{Speaker 2:} Yes, please. \newline \textbf{Speaker 1:} Now, I've finished . How do you like it? & 2, 4, 5, 5, 5 \\\hline
Opinion & \textbf{Speaker 1:} Tony looks very handsome in the suit.\newline \hspace*{0.5cm} \textbf{Speaker 2:} He prefers suits to jackets. \newline \textbf{Speaker 1:} Judge from his look, he's a very serious person. & 3, 3, 3, 4, 5 \\\hline
Non-Opinion & \textbf{Speaker 1:} Hello! Let me introduce myself . My name is Nancy. \newline \hspace*{0.5cm} \textbf{Speaker 2:} Nice to meet you, I'am Simon . I don't think I' Ve seen you around before. \newline \textbf{Speaker 1:} No, I just started working here at IBM, I am in the sale department. & 3, 4, 4, 4, 5 \\\hline
Closing & \textbf{Speaker 1:} It's hard to say . Weather is so changeable in the summer . Please listen to the latest announcement about your flight. \newline \hspace*{0.5cm} \textbf{Speaker 2:} Yes, I will . Thanks a lot . Bye! & 1, 1, 1, 2, 5 \\\hline
\end{tabular}
\caption{Example conversations from tasks with corresponding Likert ratings from writers.}
\label{tab:likert_full_convos}
\end{table*}

\section{Likert Results Compared to DailyDialog++:}\label{app:likert_metric}
We examine whether participant \edivshortspace Likert ratings are correlated to the NLI  and Sent-BERT Diversity scores from DailyDialog++ responses.  We represent a conversation's Likert rating by averaging the ratings of the 5 participants who encountered the conversation.

\begin{figure}[]
    \centering
    \begin{tabular}{c}
    \includegraphics[width=0.475\textwidth]{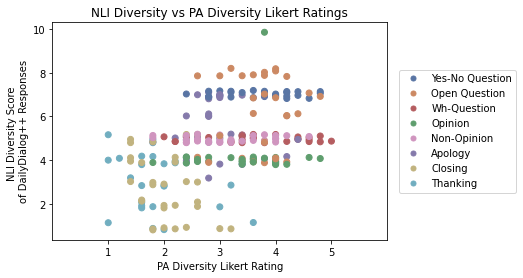}\\
    \\
    \includegraphics[width=0.475\textwidth]{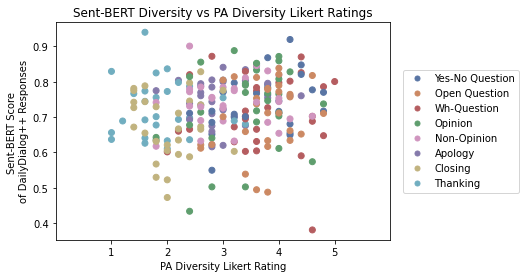}\\
    \end{tabular}
    \caption{(top) Averaged Likert ratings compared to NLI Diversity scores of DailyDialog++ responses.  Because NLI Diversity is discrete, we added a random jitter for all points between -0.2 and 0.2. 
    (bottom) Averaged Likert ratings compared to Sent-BERT Diversity scores.}
    \label{fig:likert_nli}
\end{figure}

Figure \ref{fig:likert_nli} (top) plots averaged participant Likert ratings against NLI Diversity scores for corresponding response sets in DailyDialog++.  Note that conversations were selected using median values from each speech act category; therefore, this is not a uniform sample of conversations from DailyDialog++ speech acts.  The Spearman's correlation is 0.43 (p$<$0.01), indicating participant ratings are moderately positively correlated with  NLI Diversity scores on the DailyDialog++ data.

A corresponding plot for Sent-BERT diversity can be seen in Figure \ref{fig:likert_nli} (bottom).  The Spearman's correlation is 0.19 (very weak, p$<$0.01), indicating a very weak positive correlation between  average Likert ratings and Sent-BERT diversity scores. 
Future work could investigate if a different conversation sampling strategy would yield a stronger correlation with 
Sent-BERT diversity scores.

\end{document}